\documentclass{article}

    \PassOptionsToPackage{numbers, compress}{natbib}

\usepackage[preprint]{neurips_2023}

\usepackage[utf8]{inputenc} 
\usepackage[T1]{fontenc}    
\usepackage{hyperref}       
\usepackage{url}            
\usepackage{booktabs}       
\usepackage{amsfonts}       
\usepackage{nicefrac}       
\usepackage{microtype}      
\usepackage{xcolor}         
\usepackage[pdftex]{graphicx} 
\title{Uncertainty Quantification in Deep Neural Networks through Statistical Inference on Latent Space}

\author{
  Luigi Sbail\`o \\
  Physics Department and IRIS Adlershof of the Humboldt-Universität zu Berlin\\
  Berlin, Germany\\
  \texttt{sbailo@physik.hu-berlin.de} \\
   \And
  Luca M. Ghiringhelli \\
  Physics Department and IRIS Adlershof of the Humboldt-Universität zu Berlin\\
  Berlin, Germany\\
  \texttt{luca.ghiringhelli@physik.hu-berlin.de} \\
}

\begin{document}

\maketitle

\begin{abstract}
Uncertainty-quantification methods are applied to estimate the confidence of deep-neural-networks classifiers over their predictions. However, most widely used methods are known to be overconfident. We address this problem by developing an algorithm that exploits the  latent-space representation of data points fed into the network, to assess the accuracy of their prediction. Using the latent-space representation generated by the fraction of  training set that the network classifies correctly, we build a statistical model that is able to capture the likelihood of a given prediction. We show on a synthetic dataset that commonly used methods are mostly overconfident.  Overconfidence occurs also for predictions made on data points that are outside the distribution that generated the training data. In contrast, our method can detect such out-of-distribution data points as inaccurately predicted, thus aiding in the automatic detection of outliers. 

\end{abstract}

\section{Introduction}


For the practical, widespread use of artificial-intelligence (AI) predictive models, especially where predictions would impact individuals and societies, the understanding of the limits of applicability of such models is of fundamental importance. 
An ideal AI model should be able to always provide an accurate prediction in "familiar situations", while, in "less familiar situations", it should be able to at least signal that its prediction might be inaccurate. 
The challenge is to detect such "(less) familiar situations", by using only the trained model, and therefore the training data, and the (unlabelled) test data points. This challenge is generically referred to as "uncertainty quantification" (UQ)\cite{abdar2021review}. 

Uncertainty quantification is important in applications where the consequences of incorrect predictions are severe, such as in medical imaging for diagnosing diseases \cite{li2014medical,lee2017deep,anwar2018medical,kohl2018probabilistic}, 
in autonomous-vehicles driving, where incorrect predictions can lead to accidents \cite{levinson2011towards,kalra2016driving,blum2019fishyscapes,rottmann2019uncertainty}. 
UQ is also important in scientific applications, such as physics data analysis, where neural networks with reliable estimates of prediction uncertainty and robustness are needed \cite{gal2022bayesian}. Finally, UQ is of fundamental importance in scientific and technological practice, where experiment design is performed\cite{smith2013uncertainty}. Therein, active-learning techniques \cite{gomez2018automatic} are necessarily based on reliable UQ.

Two types of uncertainties related to the data can be identified \cite{abdar2021review}, one is related to the data that are available, one with the data that are not (yet) available, i.e., it is a knowledge uncertainty.
The former is named aleatoric or irreducible uncertainty and is caused by intrinsic noise in the measurements of the data (both the descriptor and the target quantity) as well as the incompleteness of the descriptor (so that different data points are mapped into the same input description, with different target values).
The latter is named epistemic uncertainty and is caused by lack of knowledge on a portion of data. In this context, quantifying the epistemic uncertainty would be related to identifying (test) data points that are different from the training data, i.e., they belong to input domains that are distant or even disjoint from the input domain(s) of the training data.

In this paper, we focus on the estimate of the epistemic uncertainty, in particular for deep-neural-network classifiers model class. Specifically, we address the challenge of identifying data points that are "out of distribution". 
Out-of-distribution data points could be outliers or novelties/discoveries. Outliers include wrongly collected or labeled data points, typically to be discarded, while novelties/discoveries are data points in which the trained model is still applicable but belong to input domains that are distant/disjoint from the input domain(s) of the training data. In either case, the prediction of the model would require a comparison with the ground truth (assuming it is feasible) in order to decide how to proceed. The actual protocol would be related to the chosen active-learning \cite{aggarwal2014active,nguyen2019epistemic} strategy, but as a first, necessary step, we need to reliably estimate if a test point is out of distribution.

We propose a novel approach based on the analysis of the latent-space representation, which does not require a fine tuning and which we show to be both more reliable and cost-effective than the methods known to us.

\subsection{Related work}
Two broad classes of (epistemic) UQ strategies have been proposed and developed in the recent years, Bayesian techniques, including Monte Carlo (MC) based approximations (e.g., {\em MC-dropout}) or {\em ensemble} based.

\paragraph{Bayesian techniques and Monte Carlo dropout}
In an ideal Bayesian neural network, distributions of the training weights are learned, rather than specific values \cite{gal2016dropout}. 
This allows for a direct estimate of the prediction uncertainty. However, in practical application, the Bayesian architecture needs to be approximated, as it is often difficult, if possible at all to compute the exact posterior distribution. A popular approximation is the {\em MC-dropout} technique, where in production the network is run several times and a fraction of nodes are randomly switched off. This yields a distribution of prediction from which the average and standard deviation are used as actual prediction and uncertainty estimate, respectively.
It is unclear if the output distribution is always a good approximation of the posterior distribution and , in addition, results are sensitive to the dropout ratio.

\paragraph{Ensemble techniques}
{\em ensemble} techniques in general require the training of several instances of a neural network, by varying the network architecture and/or the initial seeds for the training \cite{lakshminarayanan2017simple,zhang2019active}.
However, {\em ensemble} techniques for UQ do not necessarily yield a Bayesian uncertainty estimate. In fact, {\em ensemble} techniques, also beyond the neural-network model class (e.g. random forests), are more robust when the "true model" is not part of the ensemble, which is at odds with a Bayesian estimate \cite{abdar2021review}. So, in general, it is not evident why and when an ensemble of NNs can generate good uncertainty estimates.
{\em Ensemble} techniques require training and then running of several models for prediction, which is computationally costly. 


\section{Method}
\renewcommand{\arraystretch}{1.3}

\begin{table}
\caption{Algorithm of the \textit{inference} method.}

  \label{algorithm}
  \centering
  \begin{tabular}{l}
    \toprule

    \midrule
    \textbf{Preparation:} \\
    
    - Collect input data set $\left\{X^{(0)}_i,y_i\right\}^N_{i=1}$;      \\
    - Train a feed-forward neural network to predict $f(X_i^{(0)})=y_i$;     \\
    - Create training subsets $\left\{X^{(0,k)}\right\}$ based on network prediction $k$;      \\
    - Prune all misclassified data to create confidence sets $\left\{X^{(0,k)*}\right\}$; \\
    - Collect all latent representations in confidence sets $\left\{X^{(l,k)*}\right\}$;\\
    - Compute the mean vector $\mu^{l,k}$ and covariance matrix $\Sigma^{l,k}$\\
    - Calculate the log-probability of all confidence sets based on Eq. \ref{eq:normal}. \\
    - Select the $\alpha$ and $\beta$ confidence values;\\
    - Find the $q^{(l,k)}_{\alpha}$ and $q^{(l,k)}_{\beta}$ percentiles;\\
    - Select acceptance value $a$. \\
    \textbf{Evaluation:} \\
    - Given an input point $X_i$\\
    - Predict label $f(X_i)=k$ and store latent representation $\left\{X_i^{(l,k)}\right\}$; \\
    - For each $l$ compute likelihood with  Eq. \ref{eq:smoothstep}; \\
    - Compute confidence with Eq. \ref{eq:acceptance};\\
    - If likelihood < $a$ : reject prediction; if likelihood>$a$ accept prediction.\\
    \bottomrule
  \end{tabular}

\end{table}

In the context of a data set $\left\{X_i^{(0)},y_i\right\}_{i=1}^N$, where $X_i^{(0)}\in\mathbb{R}^D$ is a $D$-dimensional input vector and $y_i\in\left\{1,\dots K\right\}$ is a class label, a deep neural network can be trained to predict the label $y_i$ from the input $X_i^{(0)}$. 
A feed-forward deep neural network is composed of $L$ consecutive layers, where each layer takes the output of the previous layer as input. The $l$-th layer, where $l\in 1\dots L$, takes an input vector $X_i^{(l-1)}$ and returns an output vector $X_i^{(l)}$ defined as:

\begin{equation}
X_i^{(l)}=\sigma\left(\sum_{i=1}^{h_L}W_{l}X_i^{(l-1)} + B_l\right),
\end{equation}

where $W_{l}$ and $b_l$ are respectively the weights matrix and bias vector learned during training, and $\sigma(\cdot)$ is a non-linear activation function. The final layer of the network returns a $K$-dimensional output vector $X^{(L)}$, and the predicted category can be inferred using the softmax function, which normalizes the output vector to obtain a probability distribution over the $K$ classes.
The latent representation of an input vector is the set of $L-1$ vectors $\left\{X^{(l)}\right\}_{l=1}^{L-1}$ that are generated when the input vector $X^{(0)}$ is fed into the network. In our method, we use this latent representation to assess the confidence of the network's prediction $f(X^{(0)})$.

The latent representations for a training set $\left\{X^{(0)}\right\}$ are obtained by forwarding the entire set through the network, which results in a set of latent representations $\left\{X_i^{(l)}\right\}_{l=1}^{L-1}$ for each input vector $X_i^{(0)}$. These latent representations are then grouped into $K$ subsets $\left\{X_i^{(l,k)}\right\}$ based on the network's predicted label $k$. In each subset, we prune all vectors $X_j$ whose predicted label $f(X_j^{(0)})$ does not match the true label $y_j$. This results in the creation of latent confidence sets $\left\{X^{(l,k)*}\right\}$, which contain the latent representations of only those vectors in the training set that the network classifies accurately. We then use these latent confidence sets to build a statistical model that can be used to estimate the likelihood of a certain prediction. The underlying assumption is that the latent representation of a well-classified input vector will be similar to the latent representations in the corresponding latent confidence set, whereas the latent representation of a misclassified input vector will be dissimilar. In essence, this approach allows us to obtain a measure of confidence for the network's predictions by leveraging the information contained in the latent representations of accurately classified training set vectors.

Every hidden layer $l$ and each class $k$ within the latent confidence set $\left\{X^{(l,k)*}\right\}$ generates an independent statistical model. For our purposes, we opt for a multivariate Gaussian distribution as the prior distribution due to its numerous advantages. 
Specifically, it is computationally efficient, scalable in high dimensions, and exhibits fast exponential decay. These properties make it an ideal choice for accurately detecting out-of-distribution points. To build the multivariate normal distribution from the latent confidence set $\left\{X^{(l,k)*}\right\}$, we compute the mean vector $\mu^{(l,k)*}$ and covariance matrix $\Sigma^{(l,k)*}$. The distribution is given by:

\begin{equation}
p(X_i^{(l,k)})=\frac{1}{\sqrt{(2\pi)^2\det\Sigma^{(l,k)*}}}\exp\left( -\frac{1}{2}(X_i^{(l,k)}-\mu^{(l,k)*})^T\Sigma^{(l,k)*}(X_i^{(l,k)}-\mu^{(l,k)*})\right).
\label{eq:normal}
\end{equation}

To assess the confidence of a prediction using a probabilistic model, we aim for high probabilities for data points where we know the model makes correct predictions and low probabilities for those with lower likelihood. 
In practice, we normalize the probability distribution in Eq. \ref{eq:normal} so that it approaches one for data points with similar likelihoods to those in the latent confidence set, and approaches zero for those with much lower likelihoods. 
To achieve this, we construct a histogram of the log-probabilities $\left\{\log p\left(X^{(l,k)*}\right)\right\}$ of the confidence sets and use it to identify the $\alpha$-th percentile $q_{\alpha}^{l,k}$ and the $\beta$-th percentile $q_{\beta}^{l,k}$. Log-probabilities below or equal to $q_\alpha^{l,k}$ are assigned a confidence value of zero, while those above or equal to $q_\beta^{l,k}$ are assigned a confidence value of one. Log-probabilities within the $q_\alpha^{l,k}-q_\beta^{l,k}$ interval are assigned a confidence value using the following smoothstep function

\begin{equation}
s(X_i)=\frac{1}{2}\left(\tanh\left(\frac{X_{i,q}-1}{2\sqrt{X_{i,q}(1-X_{i,q})}}\right)+1\right),
\label{eq:smoothstep}
\end{equation}

where $X_{i,q}=\frac{X_i-q_\alpha^{l,k}}{q_\beta^{l,k}-q_\alpha^{l,k}}$.
We notice that the function in Eq. \ref{eq:smoothstep} smoothly maps all values in the range $[q_\alpha^{l,k},q_\beta^{l,k}]$ to the interval $[0,1]$.
 To sum up, the latent confidence set $\left\{X^{l,k}\right\}^{*}$ is used to build the probability distribution in Eq. \ref{eq:normal}, which in turn is used to rank in a histogram the vectors of the latent confidence set according to their probability. The $\alpha$-th percentile $q_{\alpha}^{l,k}$ and the $\beta$-th percentile $q_{\beta}^{l,k}$ of the histogram are used in Eq. \ref{eq:smoothstep} to give the confidence of the network prediction. 

When training a deep neural network with $L$ hidden layers on a dataset with $K$ different class labels, the network generates $K*L$ probability distributions and smoothstep functions, as in Eq. \ref{eq:normal} and Eq. \ref{eq:smoothstep}. To evaluate the confidence of the network's prediction for a given input vector $X_i$ and predicted class label $k_i$, we calculate the confidence value for each hidden layer, using the activation function corresponding to the predicted class label. Specifically, we compute $L$ confidence values $s^{0,l}(X_i^{0,k}),\dots s^{0,l}(X_i^{L,k})$, where $k=y_i$.
Since each layer in the network has a large number of parameters and involves complex, nonlinear transformations in the latent space, we assume that the probability distribution of the latent representations of the same input vector but in different layers are statistically independent, i.e., $p(X_i^{l,k},X_i^{m,k})=p(X_i^{l,k})p(X_i^{m,k})$. To obtain a final confidence value, we multiply the confidence values of each hidden layer, resulting in the product 

\begin{equation}
    a(X_i^{(0,k)})=s^1(X_i^{(1,k)})*\dots*s^L(X_i^{(L-1,k)}).
    \label{eq:acceptance}
\end{equation}

 This final confidence value represents the overall confidence of the network's prediction for the given input vector $X_i^{(0,k)}$.
A summary of the algorithm is provided in Table \ref{algorithm}.

\renewcommand{\arraystretch}{1.7}
\begin{table}
\begin{center}
\begin{tabular}{c c c c c c} 
 \hline
 \textbf{Method} & \textbf{Network} & \textbf{Dropout} & \textbf{TP} & \textbf{TN} & \textbf{TN-OOD}\\ [.5ex] 
 \hline\hline
 Inference - $q_1$ & $2\cdot[1024]$  & 0.2 & $0.929\pm0.002$  & $0.732\pm0.05$ & $0.761\pm0.158$\\ 
 \hline
  Inference - $q_1$ & $2\cdot[1024]$  & 0.5 & $0.933\pm0.003$  & $0.733\pm0.046$ & $0.756\pm0.161$\\ 
 \hline
 Inference - $q_4$ & $4\cdot[256]$  & 0.1 & $0.928\pm0.003$  & $0.723\pm0.021$ & $0.707\pm0.144$\\ 
 \hline
  Inference - $q_4$ & $4\cdot[256]$  & 0.25 & $0.928\pm0.003$  & $0.745\pm0.017$ & $0.732\pm0.138$\\ 
 \hline\hline
  Inference - $q_2$ & $2\cdot[1024]$  & 0.2 & $0.873\pm0.002$  & $0.855\pm0.031$ & $0.877\pm0.1$\\ 
 \hline
  Inference - $q_2$ & $2\cdot[1024]$  & 0.5 & $0.877\pm0.004$  & $0.866\pm0.04$ & $0.879\pm0.098$\\ 
 \hline
  Inference - $q_5$ & $4\cdot[256]$  & 0.1 & $0.87\pm0.006$  & $0.858\pm0.021$ & $0.849\pm0.113$\\ 
 \hline
  Inference - $q_5$ & $4\cdot[256]$  & 0.25 & $0.869\pm0.004$  & $0.873\pm0.017$ & $0.869\pm0.103$\\ 
 \hline\hline
 Inference - $q_3$ & $2\cdot[1024]$  & 0.2 & $0.751\pm0.005$  & $0.957\pm0.014$ & $0.97\pm0.035$\\ 
 \hline
 Inference - $q_3$ & $2\cdot[1024]$  & 0.5 & $0.753\pm0.006$  & $0.968\pm0.014$ & $0.973\pm0.031$\\ 
 \hline
 Inference - $q_6$ & $4\cdot[256]$  & 0.1 & $0.748\pm0.081$  & $0.957\pm0.013$ & $0.951\pm0.058$\\ 
 \hline
  Inference - $q_6$ & $4\cdot[256]$  & 0.25 & $0.747\pm0.007$  & $0.967\pm0.012$ & $0.955\pm0.052$\\ 
 \hline\hline
 MC - dropout & $2\cdot[1024]$  & 0.2 & $0.98\pm0.002$  & $0.396\pm0.038$ & $0.267\pm0.079$\\ 
 \hline
 MC - dropout & $2\cdot[1024]$  & 0.5 & $0.902\pm0.013$  & $0.819\pm0.038$ & $0.674\pm0.125$\\
 \hline
 MC - dropout & $4\cdot[256]$  & 0.1 & $0.96\pm0.004$  & $0.564\pm0.034$ & $0.409\pm0.111$\\ 
 \hline
 MC - dropout & $4\cdot[256]$  & 0.25 & $0.908\pm0.006$  & $0.742\pm0.024$ & $0.591\pm0.141$\\
 \hline \hline
  Ensemble & $2\cdot[1024]$  & 0.2 & $0.952\pm0.009$  & $0.684\pm0.042$ & $0.517\pm0.119$\\ 
 \hline
 Ensemble & $2\cdot[1024]$  & 0.5 & $0.969\pm0.004$  & $0.594\pm0.022$ & $0.431\pm0.119$\\ 
 \hline
 Ensemble & $4\cdot[256]$  & 0.1 & $0.956\pm0.002$  & $0.67\pm0.018$ & $0.532\pm0.134$\\ 
 \hline
 Ensemble & $4\cdot[256]$  & 0.25 & $0.964\pm0.002$  & $0.621\pm0.03$ & $0.484\pm0.124$\\ 
 \hline
 \hline
\end{tabular}
\end{center}
\label{results}
\caption{Rate of true positives (TP), true negatives (TN) and true negatives on out-of-distribution samples (TN-OOD) detected using our {\em inference} method, the {\em MC-dropout} method, and the {\em ensemble} method. In the {\em inference} method the following percentiles $q=(\alpha,\beta)$ were used: $q_1=(0.01,1),q_2=(0.1,50),q_3=(3,90)$,$q_4=(2,10),q_5=(3,50),q_6=(7,90)$ Two different network architectures were employed: one composed of 2 hidden layers with 1024 nodes each - $2\cdot[1024]$,  one composed of 4 hidden layers with 256 nodes each - $4\cdot[256]$.  Dropout was implemented in every layer of all network architectures during training, with a variable dropout rate.}

\end{table}

\section{Experiments}

\begin{figure}[ht]
\centering
\includegraphics[width=1\textwidth]{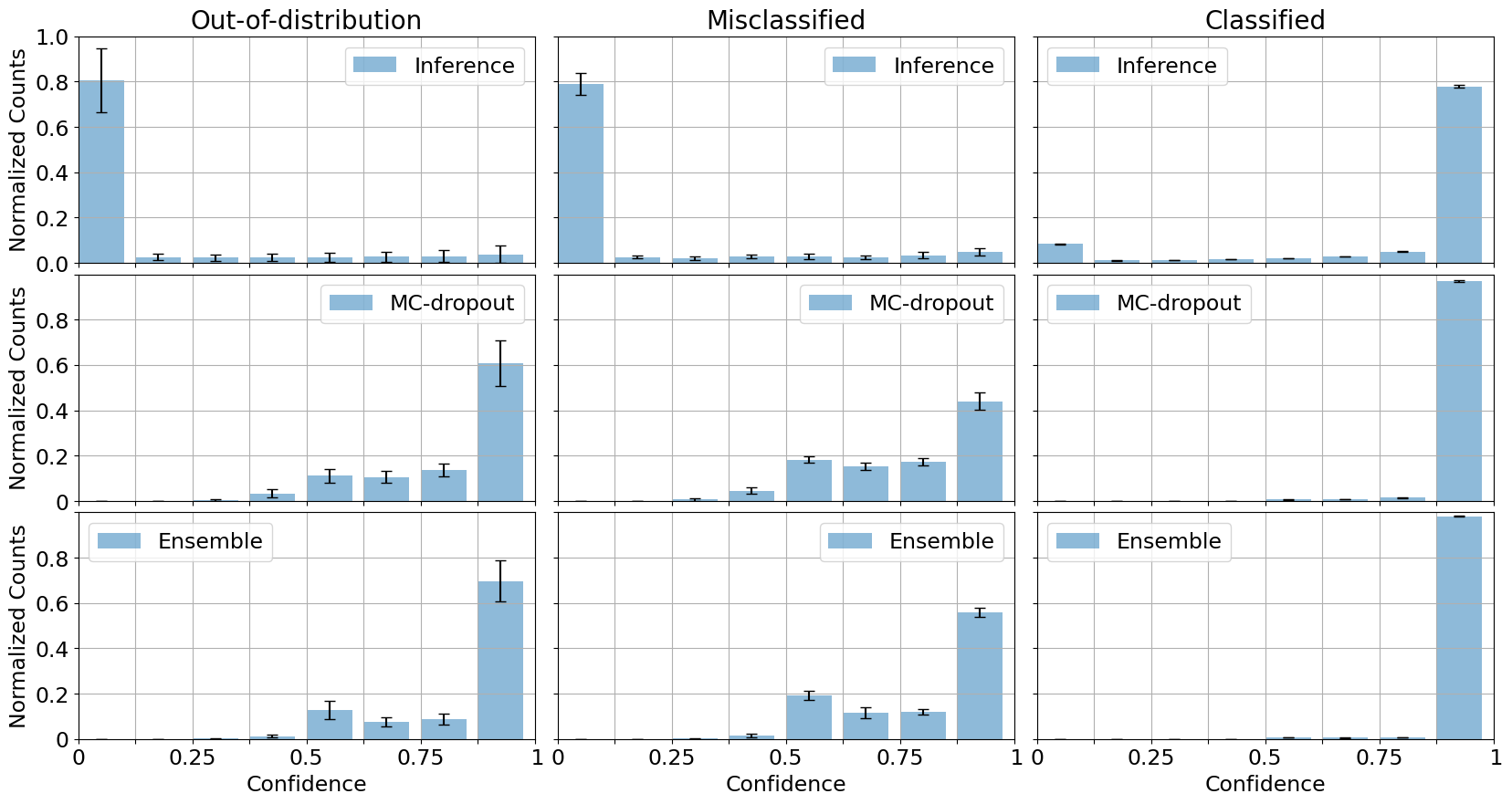}
\caption{Histograms of the confidence rate assessed on a hold out test set with the \textit{inference}, the 'MC-dropout' and 'ensemble' methods. The histogram is relative to the experiment conducted with a network composed of 2 hidden layers with 1024 nodes each and dropout rate fixed to 0.5. The \textit{inference} method employs percentile values $q_2=(0.1,50)$. We can see in the histograms that most \textit{misclassified} and {\em out-of-distribution} samples are assigned a confidence close to zero with the \textit{inference} method, while the \textit{MC-dropout} and the \textit{ensemble} methods assign much higher values to the same samples. Note that, in this figure, we show the histograms relative to the experiment where the \textit{MC-dropout} method features the best values in detecting true negatives. }
\label{histograms}
\end{figure}

To demonstrate the validity of our method, we conducted numerical experiments and compared its performance with state-of-the-art methods using the MNIST dataset. Specifically, we trained feedforward neural networks with varying numbers of layers to classify the images. To showcase the ability of our method to detect out-of-distribution samples, we conducted experiments where we removed all samples from the training set associated with a specific label, and then trained the network on the remaining set. This allowed us to evaluate the network's capability to identify instances that it has never encountered during training, and to determine whether it can recognize when it does not know the answer.

We utilized two distinct network architectures for our experiments. The first architecture was shallower and comprised of two layers, with each layer having 1024 nodes. The second architecture was deeper, consisting of four layers with 256 nodes each. We employed early stopping during training, which terminated the process once an accuracy threshold of 0.96 was attained. In all training we used dropout with different values.
To evaluate methods performance, we used a hold-out test set that excluded out-of-distribution samples and thus sampled from the same distribution as the training set, as well as a set consisting only of out-of-distribution samples. 
We conducted the experiment iteratively by removing each label one-by-one and evaluating the confidence of the network's predictions using our inference method, MC dropout, and {\em ensemble} methods.
The MC dropout method involves obtaining an ensemble of predictions by leaving dropout activated during prediction. In contrast, the {\em ensemble} method involves training models with different initial conditions to obtain an ensemble of predictions. Each experiment took approximately 25 minutes for the {\em inference} and {\em MC-dropout} methods, and approximately 4 hours for the {\em ensemble} method.
All simulations were performed on a single CPU Intel(R) Core(TM) i7.
To make a prediction using an {\em ensemble}, we take the most frequently predicted value as the network prediction. The uncertainty associated with this prediction is determined by the fraction of predictions that match the most frequently predicted value, relative to the total number of models used in the ensemble. We used 100 passes for the {\em MC dropout} (with ratio equal to the one used in training) experiments and 10 models for the {\em ensemble} experiments. Uncertainty in our method is calculated using the algorithm described in Table \ref{algorithm}.

The objective of our experiments is to compare the aforementioned uncertainty quantification methods and assess the algorithms' ability to differentiate between well-classified and misclassified predictions. To achieve this, we calculate uncertainties on separate holdout test sets and reject predictions whose uncertainty falls below a certain threshold, denoted as $a$.
We then assess the method's performance by counting the number of well-classified vectors that are assigned a value above the threshold, which represents true positives (TP). 
Additionally, we count the number of misclassified vectors below the threshold (TN) and the number of vectors from the out-of-distribution test set below the threshold (TN-OOD). These three values serve as indicators to evaluate the quality of the method, with higher values indicating better performance.
Of particular importance in our analysis is the TN-OOD value, as it helps us understand the algorithm's ability to identify samples that lie outside the distribution of the training set.
For the {\em MC-dropout} and {\em ensemble} methods, we employ a high threshold of $a=0.99$ due to the tendency of both these methods to exhibit overconfidence\cite{abdar2021review}. Conversely, our inference method employs the conceptually intuitive threshold of $a=0.5$.
Our method utilizes two parameters: the percentiles $\alpha$ and $\beta$. These parameters determine the distribution of confidence values within the sets of vectors. Specifically, the percentage of vectors assigned a confidence value of 1 is equal to $100-\beta$, while the percentage of vectors assigned a value of 0 is equal to $\alpha$.
The parameters $q=(\alpha,\beta)$ used in our experiments are: $q_1=(0.01,1),\ q_2=(0.1,50),\ q_3=(3,90)$ in the network composed of 2 hidden layers with 1024 nodes each, $q_4=(2,10),\ q_5=(3,50),\ q_6=(7,90)$ in the network composed of 4 hidden layers with 256 nodes each.

\section{Discussion}

The results of our experiments are summarized in Table \ref{results}. The table demonstrates that our {\em inference} method successfully detects both high true positives and true negatives. By tuning the percentiles $q=(\alpha,\beta)$, we can determine the balance between true positive and true negative detection.
At the top of the table, using values $q_1$ and $q_2$ yields high true positives. Using values $q_2$ and $q_5$ provides a balance between true positives and true negatives, while values $q_3$ and $q_6$ below allow for high true negatives. 
Our method consistently produces reliable results, regardless of the dropout rate and number of hidden layers. 
This is achieved by tuning the percentiles while utilizing different network architectures.
As previously discussed, both the {\em MC-dropout} and {\em ensemble} methods demonstrate overconfidence in their predictions. They exhibit high true positive detection and low true negative detection. As expected, {\em MC-dropout} shows higher true negatives and lower true positives with increased dropout rates. Conversely, the {\em ensemble} method seems to follow the opposite trend.

Our method demonstrates robust performance in detecting true negatives, even in the presence of out-of-distribution samples, unlike the {\em MC-dropout} and {\em ensemble} methods, as we can see in Fig. \ref{histograms}.
In our experiments, the {\em MC-dropout} method achieves a maximum out-of-distribution sample detection rate of only 68\%. Similarly, the {\em ensemble} method reaches a maximum of 53\% true negative detection for out-of-distribution samples.
In contrast, our {\em inference} method consistently maintains a true negative out-of-distribution sample detection rate above 70\%, even when percentile values are adjusted to prioritize true positive detection over true negatives. Notably, our {\em inference} method achieves a true negative detection rate of over 95\% while still maintaining a reasonable level of approximately 75\% true positive detection.
However, our {\em inference} method seems to be slightly less reliable when it comes to true positive detection compared to the other methods, although the difference is minimal. Our method can still achieve over 92\% true positive detection. Increasing the true positive detection rate can be achieved by tuning the percentile values, but this would come at the cost of decreasing the true negative detection rate. Since our focus is on the capability of the method to detect true negatives, we don't show higher values of true positives that would compromise this capability.
Additionally, it is possible to adjust the acceptance rate threshold, currently fixed at $a=0.5$, to further balance the detection towards true negatives or true positives. However, in these experiments, tuning the percentile values proves to be sufficient to obtain satisfactory results.

We would like to emphasize that our {\em inference} method not only yields superior overall results but also offers greater ease of application compared to other methods.
For instance, the {\em MC-dropout} method requires the network to utilize dropout, which may adversely affect network performance in many applications. Moreover, when dropout can be employed, finding the optimal dropout rate necessitates training multiple models until a suitable rate is discovered.
The ensemble method instead, while conceptually simpler to apply, demands training multiple models to assess uncertainty. This computational expense renders it unfeasible for many applications.
In contrast, our {\em inference} method only requires tuning the percentile values, a task that can be performed after training. This means that only one model needs to be trained, simplifying the process. Furthermore, the computational cost remains low as it scales quadratically with the number of nodes in the layers, which typically do not exceed thousands of units. In practice, the time required to construct the statistical model employed in our method is negligible compared to the time needed to train the neural network.
Tuning the percentile parameters can be done after training on a test set. Since our {\em inference} method demonstrates consistent results on misclassified test vectors sampled from the same distribution as the training set, as well as on out-of-distribution samples (see Table \ref{results}), parameter tuning on a test sample can effectively generalize to out-of-distribution scenarios.

\section{Limitations and outlook}

Although the {\em inference} method demonstrates promising results, it is important to acknowledge its limitations. Firstly, the method assumes that the data are well approximated by a normal distribution in the latent space. Although this assumption holds true in our experiments, it is possible that in more complex scenarios, a different prior or inference method might be more appropriate.
Additionally, the assumption of statistical independence among the probability distributions across different layers in the latent space, as shown in Eq. (\ref{eq:acceptance}), is quite strong. When dealing with numerous hidden layers, this assumption can potentially impact the quality of the results.
Nevertheless, it is worth mentioning that this method can still be utilized selectively on certain layers, where the assumption of independence among probabilities holds true.

We employed a technique to measure uncertainty in classification, which involved creating a distinct statistical model for each potential label. This approach played a significant role in our analysis.
It would be advantageous to extend this method to inference problems that involve predicting continuous values. In such scenarios, we cannot rely on the differentiation of data based on their labels, so an alternative strategy must be employed. One potential strategy is to utilize a statistical model based on the training set vectors, where the model's prediction is similar to the prediction we wish to assess the uncertainty for.
However, the specific measure of similarity needs to be defined carefully to ensure accurate results.

\section{Conclusions}

This paper presents a novel approach for quantifying uncertainty in classification tasks using feed-forward neural networks. We conducted a comparative analysis of our method against state-of-the-art techniques on a benchmarking synthetic dataset. The results demonstrate that our method outperforms the examined state-of-the-art approaches, particularly excelling in the detection of true negatives. This advantage extends to out-of-distribution samples as well, where other methods fall short.

Although it is difficult to assess the impact of a work in a rapidly evolving field like machine learning, we believe that this work can have a positive impact in the field.
Especially in contexts where the identification of out-of-distribution samples is crucial, such as autonomous driving and active learning.

\section*{Acknowledgements}
LS acknowledges for funding the Leibniz Association, project "Memristors Materials by Design, MeMabyDe". LMG acknowledges for funding the NFDI project "FAIRmat" (FAIR Data Infrastructure for Condensed-Matter Physics and the Chemical Physics of Solids, German Research Foundation, project Nº 460197019).

\bibliography{bib}

\end{document}